\documentclass[review]{elsarticle}
\usepackage{amsmath}
\usepackage{amssymb}
\usepackage{algorithm,algorithmic}
\usepackage{tabularx}
\usepackage{booktabs}
\usepackage{lineno,hyperref}
\usepackage{subfig}

\usepackage{color}
\pdfoutput=1
\journal{Journal of LaTex template}
\biboptions{authoryear}

\begin{document}
\author{Chiranjibi Sitaula \corref{cor1}}
\cortext[cor1]{Corresponding Author}
\ead{csitaul@deakin.edu.au}
\author{Yong Xiang}
\author{Sunil Aryal}
\author{Xuequan Lu}
\begin{frontmatter}

\title{Scene Image Representation by Foreground, Background and Hybrid Features}

\address{School of Information Technology, Deakin University, Geelong, Victoria 3216, Australia}

\begin{abstract}
Previous methods for representing scene images based on deep learning primarily consider 
either the foreground or background information
as the discriminating clues for the classification task. However, scene images also require additional information (hybrid) to cope with the inter-class similarity and intra-class variation problems. In this paper, we propose to use hybrid features in addition to foreground and background features to represent scene images. We suppose that these three types of information could jointly help to represent scene image more accurately. To this end, we adopt three VGG-16 architectures pre-trained on ImageNet, Places, and Hybrid (both ImageNet and Places) datasets for the corresponding extraction of foreground, background and hybrid information. All these three types of deep features are further aggregated to achieve our final features for the representation of scene images. Extensive experiments on two large benchmark scene 
datasets (MIT-67 and SUN-397) show that our method produces the state-of-the-art classification performance. 
\end{abstract}


\end{frontmatter}


\section{Introduction}
With the prevalent and rising demand of robotics and video surveillance, image representation has been a very important \citep{sitaula2019indoor} field to improve classification and recognition accuracies
. However, the image representation depends on the problem domain, because we need to represent the images according to contents present in the images and all images can hardly be represented by a single features extraction method. Broadly, image representation methods are categorized into two categories: content-based image representation methods and context-based image representation methods.

Content-based image representation methods \citep{zeglazi_sift_2016, oliva2005gist,oliva_modeling_2001,dalal2005histograms,lazebnik2006beyond,wu_centrist:_2011, xiao_mcentrist:_2014,margolin2014otc,quattoni_recognizing_2009,zhu_large_2010,li2010object,parizi2012reconfigurable,juneja2013blocks,lin_learning_2014,shenghuagao2010local,perronnin2010improving,xiao2010sun,sanchez2013image,gong_multi-scale_2014,kuzborskij2016naive,he2016deep,zhou2016places,zhang2017image,tang_g-ms2f:_2017,8085139,guo2016bag,bai2019coordinate,zhou2014learning,simonyan2014very,yang2015multi,wu2015harvesting,dixit2015scene} rely on the visual content information of the scene images. These features are either based on traditional computer vision based algorithms \citep{zeglazi_sift_2016,oliva2005gist,oliva_modeling_2001, dalal2005histograms,lazebnik2006beyond, wu_centrist:_2011, xiao_mcentrist:_2014,margolin2014otc,quattoni_recognizing_2009,zhu_large_2010,li2010object,parizi2012reconfigurable,juneja2013blocks,lin_learning_2014,shenghuagao2010local,perronnin2010improving,xiao2010sun,sanchez2013image} or deep learning based algorithms \citep{gong_multi-scale_2014, kuzborskij2016naive,he2016deep,zhou2016places, zhang2017image,tang_g-ms2f:_2017,8085139,guo2016bag,bai2019coordinate,zhou2014learning,simonyan2014very,yang2015multi,wu2015harvesting,dixit2015scene}. Traditional computer vision based algorithms are more suitable for specific types of images such as texture images. However, the recent studies have shown that deep learning based algorithms have higher classification performance than the traditional computer vision based algorithms for the complex scene images involving objects and their associations.

\begin{figure}[tb]
\begin{center}
 \subfloat[]{\includegraphics[width=0.45\textwidth, height=27mm,keepaspectratio]{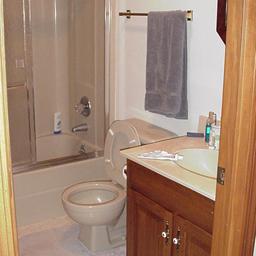}}
\hspace{1pt}
 \subfloat[]{\includegraphics[width=0.45\textwidth, height=27mm,keepaspectratio]{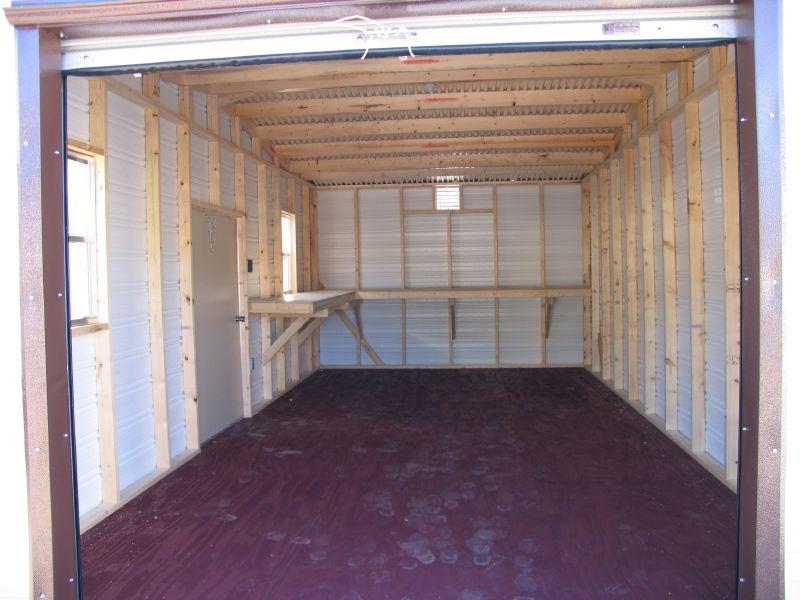}}
 \hspace{1pt}
 \subfloat[]{\includegraphics[width=0.45\textwidth, height=27mm,keepaspectratio]{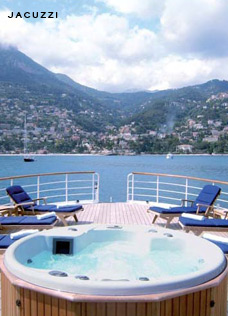}}\\
 \subfloat[]{\includegraphics[width=0.45\textwidth, height=27mm,keepaspectratio]{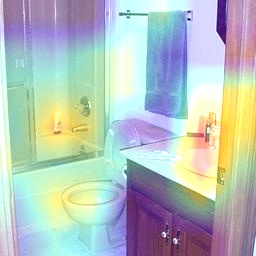}}
\hspace{1pt}
 \subfloat[]{\includegraphics[width=0.45\textwidth, height=27mm,keepaspectratio]{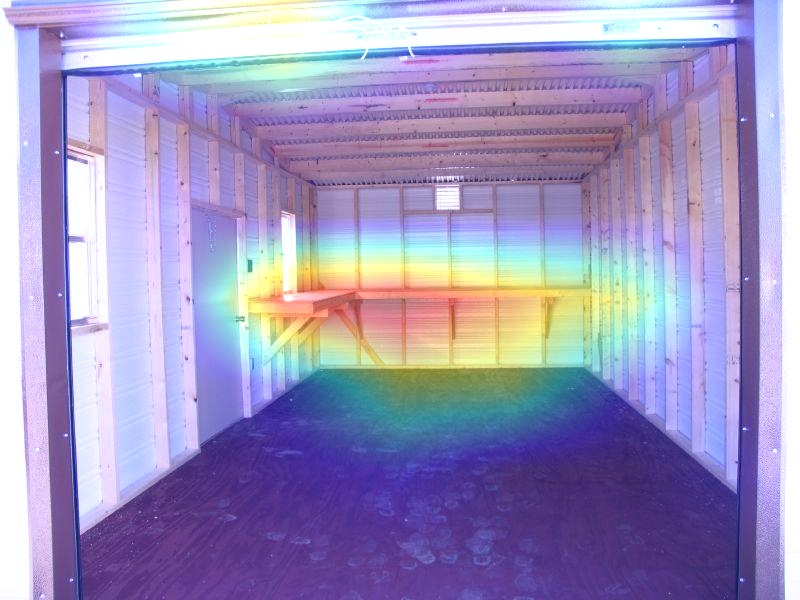}}
 \hspace{1pt}
 \subfloat[]{\includegraphics[width=0.45\textwidth, height=27mm,keepaspectratio]{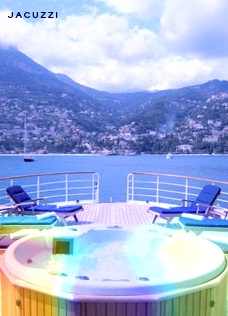}}
   \caption{ First row shows the scene images belonging to three different categories ((a) toilet, (b) garage, and (c) bathtub outdoor) and the second row shows their corresponding discriminating clues ((d) foreground information for toilet, (e) background information for garage, and (c) hybrid information for bathtub outdoor)). Note that all the feature maps are extracted from the $5^{th}$ pooling layer of VGG-16 models, which are pre-trained on ImageNet, Places, and hybrid datasets (ImageNet+Places).}
  \label{fig:1}
 \end{center}
  \end{figure}

Context-based image representation \citep{zhang2017image,wang2019task,sitaula2019tag} addresses the difficult problem of representing the ambiguous images including between-class similarity and within-class dissimilarity. These works are mostly performed based on the exploitation of human annotations/descriptions, with regard to the similar scene images of the input query image on the web. Nevertheless, web crawling and features extraction based on such approaches could be sometimes impractical due to the labor-intensive computational requirements and multiple levels of pre-processing such as tokenization of raw crawled texts, language translation, stemming and lematization, etc.  

\begin{figure}[tb]
\begin{center}
 \subfloat[Foreground]{\includegraphics[width=0.45\textwidth, height=50mm,keepaspectratio]{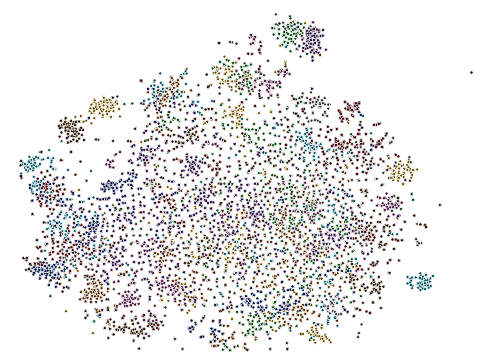}}
\hspace{0pt}
 \subfloat[Background]{\includegraphics[width=0.45\textwidth, height=50mm,keepaspectratio]{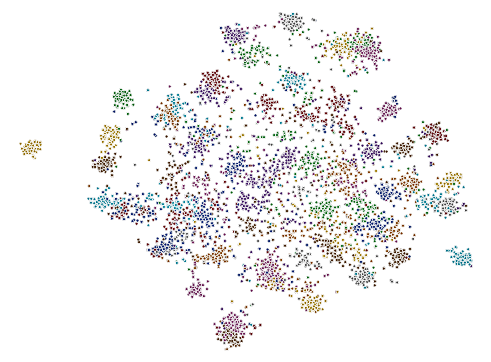}}
 \hspace{0pt}
 \subfloat[Hybrid]{\includegraphics[width=0.45\textwidth, height=50mm,keepaspectratio]{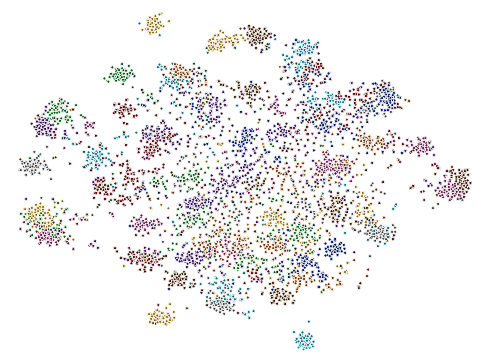}}
 \hspace{0pt}
  \subfloat[Aggregated]{\includegraphics[width=0.45\textwidth, height=50mm,keepaspectratio]{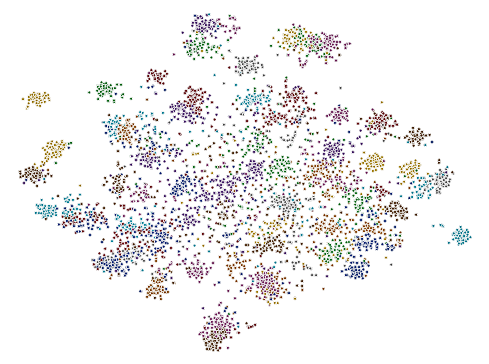}}\\
   \caption{The t-SNE visualization of scatter plots showing the discriminability of features based on the (a) foreground, (b) background, (c) hybrid, and (d) aggregated information. Note that multiple colors represent categories (67 categories) in the MIT-67 dataset.}
  \label{fig:2}
 \end{center}
  \end{figure}

The existing methods in the literature primarily focus on either foreground or background information
, which may not be sufficient for accurate representation of varying types of scene images. First, different types of scene images may require different types of information to distinguish them accurately. Fig. \ref{fig:1} shows an example.
In the figure, three types of scene images requires three different types of information for their better separability.
Second, scene images usually involve inter-class similarity and intra-class dissimilarity issues. It may require additional information (hybrid) to foreground and background information to improve the separability.

To bridge the aforementioned gaps above, we perform the fusion of three different types of information including foreground, background, and hybrid for each image. 
For this, we extract the foreground, background, and hybrid information of each image with the help of VGG-16 models \citep{simonyan2014very} pre-trained on ImageNet \citep{deng_imagenet:_2009}, Places \citep{zhou2017places}, and both (ImageNet + Places), respectively. We choose the VGG-16 model due to its simple architecture yet prominent features extraction capability \citep{guo2016bag,8085139,bai2019coordinate}. To achieve the corresponding features from the VGG-16 models, we utilize a higher level pooling layer ($p\_5$) \citep{8085139,guo2016bag} as the features extraction layer, because we found that the $p\_5$ layer yields highly separable features than other layers (see detail in Section \ref{pooling_methods})
. Finally, we aggregate these three types of features to achieve our final features for the classification. The separability of our aggregated features (combined) and individual features are shown in Fig. \ref{fig:2}, using t-SNE (t-Distributed Stochastic Neighbor Embedding) scatter plot visualization tool.

The main {\bf contributions} of this paper are summarized as follows.
\begin{enumerate}[(a)]
    \item We propose a novel method for image representation by identifying three different types of information (foreground, background, and hybrid) and fusing them. 
    \item We design an effective scheme to aggregate three important types of information using three different pre-trained deep learning models (VGG-16). We analyze five pooling layers of VGG-16 and choose the best features extractor in this work.
    \item We evaluate our method on two common scene benchmark datasets: MIT-67 \citep{quattoni_recognizing_2009} and SUN-397 \citep{xiao2010sun}. Evaluation results show that our method produces features with better separability and results in the state-of-the-art classification performance.
\end{enumerate}

The rest of the paper is organized as follows. Section \ref{lit_review} reviews the previous image representation methods for the scene images. Section \ref{prop_method} elaborates our proposed method in a step-wise manner. Section \ref{experiment} explains the implementation details, the comparisons with the previous methods and the ablative studies. Section \ref{conclusion} concludes this work.

\section{Related work}
\label{lit_review}
In general, image representations can be divided into two types: content-based and context-based.

\subsection{Content-based image representation}
Content-based image representation methods are further categorized into two subgroups: traditional computer-vision based algorithms and deep-learning based algorithms. Traditional computer vision-based algorithms primarily depend on the traditional feature extraction methods such as GIST-color \citep{oliva_modeling_2001},
Generalized Search Trees (GIST) \citep{oliva2005gist},
Histogram of Gradient (HOG) \citep{dalal2005histograms},
Spatial Pyramid Matching (SPM) \citep{lazebnik2006beyond},
RoI (regions of interest) with GIST\citep{quattoni_recognizing_2009},
MM (Max-Margin)-background\citep{zhu_large_2010},
Object bank\citep{li2010object},
Improved Fisher Vector (IFV)\citep{perronnin2010improving},
Laplacian Sparse coding SPM (LscSPM)\citep{shenghuagao2010local},
CENsus TRansform hISTogram (CENTRIST) \citep{wu_centrist:_2011},
Reconfigurable Bag of Words (RBoW)\citep{parizi2012reconfigurable},
Bag of Parts (BoP)\citep{juneja2013blocks},
multi-channel CENTRIST (mCENTRIST) \citep{xiao_mcentrist:_2014},
Important Spatial Pooling Region (ISPR)\citep{lin_learning_2014},
Oriented Texture Curves (OTC) \citep{margolin2014otc},
Scale Invariant Feature Transform (SIFT) \citep{zeglazi_sift_2016},
and so on. The feature extraction algorithms under such traditional computer vision-based algorithms emphasize the 
low-level details of the images such as colors, intensity, gradients, orientations, etc. In other words, such algorithms are mostly local details oriented, and therefore, more suitable for specific types of images such as texture images. They are usually not ideal to represent the complex types of images such as scene images. Also, the feature size extracted by such algorithms is mostly higher than recent high-level features.

Furthermore, deep learning-based methods \citep{zhang2017image,gong_multi-scale_2014,guo2016bag,8085139,tang_g-ms2f:_2017,kuzborskij2016naive,zhou2016places,he2016deep,bai2019coordinate} are found to have noticeably better classification accuracies than existing traditional methods. Recent deep learning-based algorithms for scene representations are:  CNN-MOP \citep{gong_multi-scale_2014},
CNN-sNBNL \citep{kuzborskij2016naive},
VGG \citep{zhou2016places},
ResNet152 \citep{he2016deep},
EISR \citep{zhang2017image},
G-MS2F \citep{tang_g-ms2f:_2017},
SBoSP-fusion \citep{guo2016bag},
BoSP-Pre\_gp \citep{8085139},
CNN-LSTM \citep{bai2019coordinate}, and so on.
\citet{gong_multi-scale_2014} 
and \citet{kuzborskij2016naive} employed Caffe model \citep{jia2014caffe} to achieve features from the $FC$-layer for the scene images classification purpose. Gong et al. extracted multi-scale order-less features, which were obtained by extracting the global activation features ($FC$-layer) for each scale of the images and aggregated using the Vector of Locally
Aggregated Descriptors (VLAD) pooling method. Similarly, Kuzborskij et al. also feed the output of $FC$-layers into the Naive Bayes Nearest Neighborhood classifiers.
\citet{zhou2016places} released a new places related dataset to train the popular deep learning model such as VGG model \citep{simonyan2014very}. This leads to a promising classification accuracy of the images, especially scene images.
\citet{he2016deep} proposed a novel architecture for deep learning, which followed the residual concepts and outperformed the previous off-the-shelf deep learning models such as the VGG model \citep{simonyan2014very}, GoogleNet model \citep{szegedy2015going}, etc.
\citet{zhang2017image} sliced an image into multiple sub-images using random slicing and extracted deep features for each slice. Deep features of each slice were concatenated as combined deep features of the corresponding image. Finally, the fusion of such combined deep features with tag-based features yielded the final features representing an image for the classification.
\citet{tang_g-ms2f:_2017} introduced a score-fusion technique to provide the probability-based deep features. They employed three intermediate classification layers of the GoogleNet model \citep{szegedy2015going} for the score fusion, which improved classification performance remarkably.
Guo et al. \citep{guo2016bag,8085139} adopted the VGG-16 model to extract the deep features by developing the concept of the bag of surrogate parts (BoSP).
Their method provides features with a fixed-size, which is lower than the most of the existing methods for the scene image representation despite the prominent classification accuracy.
Recently, \citet{bai2019coordinate} established a new deep learning model by incorporating Convolutional Neural Networks (CNNs) with Long Short Term Memory networks (LSTMs). They believe that the ordered slices of images as a sequence problem could be solved by the LSTMs model on top of CNNs model for the scene image classification. Their method, thus, offers prominent classification accuracy on scene images.

\subsection{Context-based image representation}
There have been some recent works \citep{zhang2017image,wang2019task,sitaula2019tag} under context-based image representation methods. The main motivation of using such features is the use of human knowledge scattered in the form of annotation/description on the web, based on which people may be able to distinguish confusing complex scene images. For this,  \citet{zhang2017image} extracted the description of the related images on the web to design the bag of words (BoW) features directly. Their method  suffers from not only the occurrence of outliers but also the curse of higher feature size. To tackle this limitation,  \citet{wang2019task} devised the concept of filter bank using the category labels of ImageNet and Places to filter out the outliers to some degree. However, their method do not filter out the outliers accurately. The main reason of it not being able to filter out outliers is the dependence on pre-defined category labels only for the filter banks. This results in retrieval of more task-generic tags which are not suitable for scene images. To address such problem raised in the previous works, \citet{sitaula2019tag} developed a novel domain-specific filter bank that extracts the tag-based features by leveraging the semantic similarity of tags with scene image category labels. Their method not only generates rich tag-based features with the use of such filter banks, but also reduces the feature size of an image significantly. 

To sum up, the existing content-based and context-based image representation methods based on deep learning models outperform previous methods in most cases.
This motivates us to explore further in content-based image representation based on the deep learning models, to achieve better representation of scene images. However, such methods in the literature have two main limitations. First, the existing methods consider either foreground or background information to represent the scene images, which may not be sufficient clue to some of the scene images requiring additional information such as hybrid information as a discriminating clue (see in Fig. \ref{fig:1}). Second, scene images may contain three different types of information including foreground, background and hybrid information, which are complementary to each other in scene image representation tasks. Thus, we propose to fuse three different types of information (foreground, background and hybrid) that describe distinguishing clues in the scene representation. Our features bolster the classification performance significantly.
\begin{figure*}[tb]
\begin{center}
 \includegraphics[width=\textwidth, height=90mm,keepaspectratio]{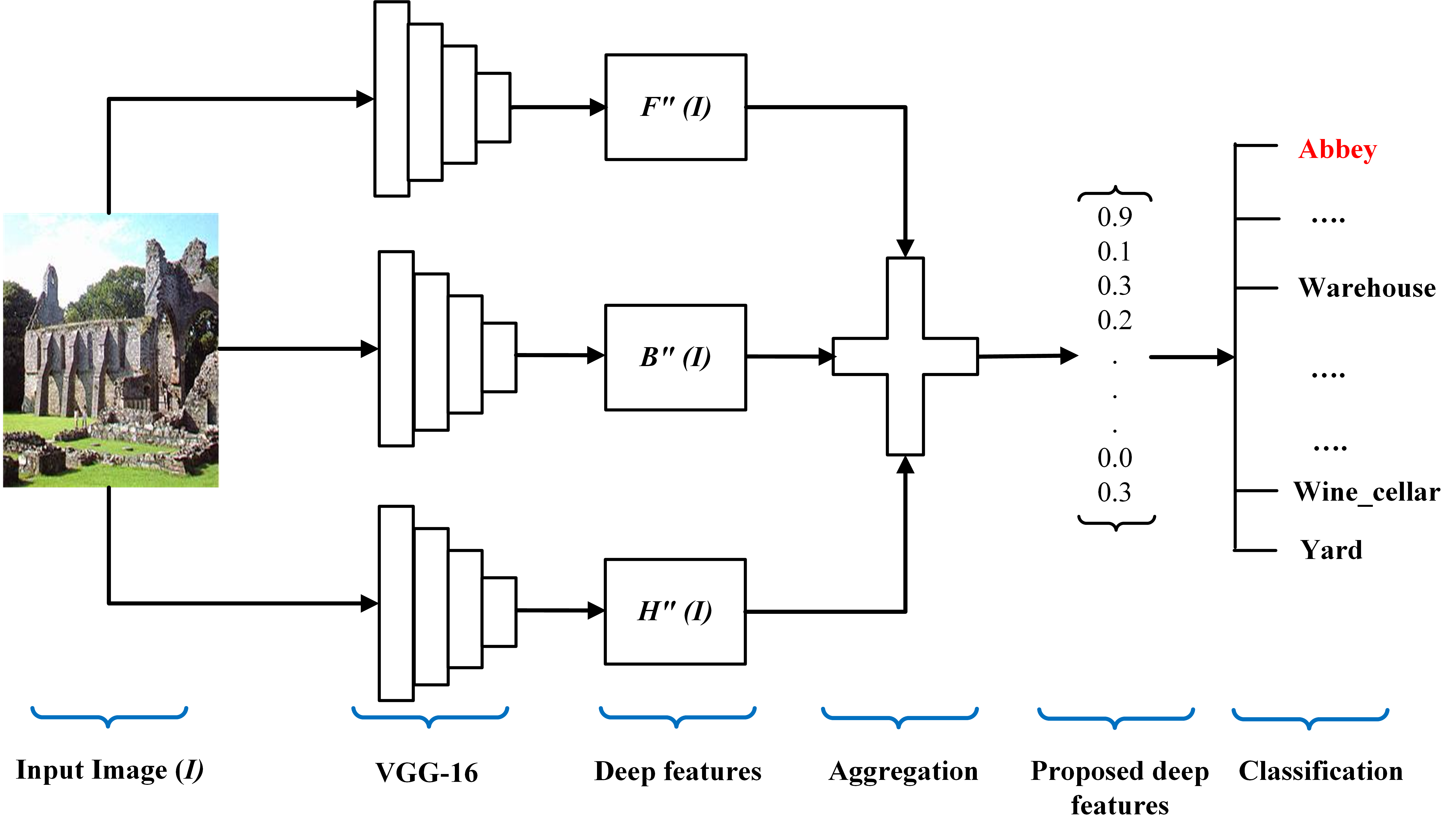}
  \caption{Overall pipeline of the proposed method. Three different pre-trained VGG-16 models yield foreground, background, and hybrid features represented by $F''(I)$, $B''(I)$ and $H''(I)$, respectively. }
  \label{fig:pipeline}
 \end{center}
  \end{figure*}

\begin{figure*}[t]
\begin{center}
 \includegraphics[width=0.90\textwidth, height=90mm,keepaspectratio]{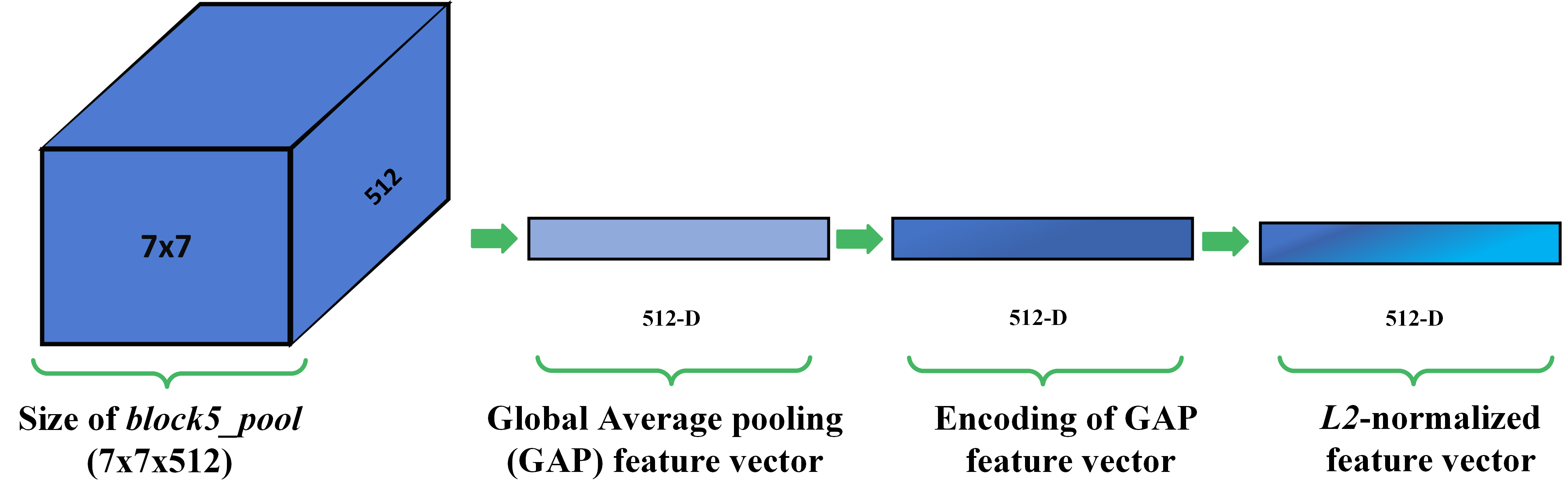}
  \caption{Three steps to achieve normalized feature vector after global average pooling (GAP) operation on the $5^{th}$ pooling layer ($p\_5$) followed by encoding step and normalization step.}
  \label{fig:gap_norm}
 \end{center}
  \end{figure*}

\section{Proposed method}
\label{prop_method}
To extract the proposed features, we follow four steps:  foreground features extraction (Section \ref{foreground_features}), background features extraction (Section \ref{background_features}),  hybrid features extraction (Section \ref{hybrid_features}), and their aggregation (Section \ref{aggregation}).
The overall pipeline of our proposed method is shown in Fig. \ref{fig:pipeline}. 
For each of the first three steps, we have (1) global average pooling (GAP) features extraction that helps to capture both higher and lower activation values appropriate for scene representation, (2) encoding and (3) normalization. It is shown in Fig. \ref{fig:gap_norm}. All the normalized feature vectors achieved are aggregated to achieve our proposed features as the final scene image representation.

\subsection{Foreground features extraction}
\label{foreground_features}
Foreground features often capture the object-based information in the scene images. These features are extracted from deep learning models pre-trained on ImageNet which consists of object images of $1,000$ categories. There are several pre-trained models such as ResNet-50 \citep{he2016deep}, GoogleNet \citep{szegedy2015going}, etc. for the foreground features extraction; nevertheless, we use VGG-16 model that has been frequently used in scene representation tasks \citep{guo2016bag,8085139,bai2019coordinate} due to its
simplicity and prominent features extraction capability.
We represent VGG-16 model pre-trained on ImageNet as $F$.
Here, Eq. \eqref{eq:1} extracts GAP (Global Average Pooling) 
features $F(I)$ from the $p\_5$ layer of the model $F$ (see details in Sec. \ref{line_graph:pooling}). For the extraction of such GAP features, we average each feature map with height ($h$) and width ($w$). Similarly, the depth ($d$) represents the total number of feature maps in the particular tensor.
For instance, the $5^{th}$ pooling layer ($p\_5$) of VGG-16 model has a three dimensional tensor of height ($h$), width ($w$), and depth ($d$) as $7$, $7$ and $512$, respectively. This results in the features size of $512$-D after GAP operation on each feature map of the tensor.
For this, we assume that the symbol $F_{j}^{i}$ represents the $i^{th}$ activation value of the $j^{th}$ feature map of the tensor. 
\begin{equation}
{F(I)}=\frac{1}{h*w}*\bigg\{\sum_{i=1}^{h*w}{F_{1}^i}, \sum_{i=1}^{h*w}{F_{2}^i}, ,\cdots \sum_{i=1}^{h*w}{F_{d}^i}\bigg\},
\label{eq:1}    
\end{equation}
GAP features ($F(I)$) achieved from Eq. \eqref{eq:1} are represented by the vector elements such as $\{f_1,f_2,\cdots f_d \}$, where $d$ is the size of such features and 
$f_j=\frac{\sum_{i=1}^{h*w}F_{j}^{i}}{h*w}$.
These features are encoded as suggested by \citet{8085139} and \citet{guo2016bag}, which has been found prominent for feature map encoding during the foreground based features extraction. Rather than utilizing such encoding for each feature map in \citet{8085139} and \citet{guo2016bag} 
, we employ it in our GAP features, which yields $F'(I)$ 
shown in Eq. \eqref{eq:2}. GAP features provide the discriminating information of scene images because it helps leverage both higher and lower activation values, which are discriminating clues of scene images classification.
\begin{equation}
{F'(I)}=\begin{cases} 
      0 & \text{if $f_j< mean(F(I))$}, \\
      \frac{f_j}{max(F(I))} & \text{if $f_j\geq mean(F(I))$}.
   \end{cases},
\label{eq:2}    
\end{equation}

The encoded features are normalized using $L2$-norm to obtain $F''(I)$ as shown in Eq. \eqref{eq:3}. While doing normalization, we add epsilon (i.e., $\epsilon=1e-7$) with the denominator to avoid the divide-by-zero exception.
\begin{equation}
{F''(I)}=\frac{F'(I)}{\| F'(I)\|_2+\epsilon},
\label{eq:3}    
\end{equation}
Eq. \eqref{eq:3} yields the foreground features ($F''(I)$) of the input image.

\subsection{Background features extraction}
\label{background_features}
Background features represent the global layout information present in the images. These features are extracted from the deep learning model pre-trained on Places that involves background images of $365$ categories. We represent VGG-16 model pre-trained on Places as $B$ in this work. The GAP features extracted from this model is shown in Eq. \eqref{eq:4}. Let $B_{j}^{i}$ represent the $i^{th}$ activation value for the corresponding $j^{th}$ feature map, and $B(I)$ represent the GAP features extracted from $B$. 

\begin{equation}
{B(I)}=\frac{1}{h*w}*\bigg\{\sum_{i=1}^{h*w}{B_{1}^i}, \sum_{i=1}^{h*w}{B_{2}^i}, ,\cdots \sum_{i=1}^{h*w}{B_{d}^i}\bigg\}.
\label{eq:4}    
\end{equation}

GAP features $B(I)$, which are represented by $\{b_1,b_2,\cdots,b_d\}$, are further encoded and normalized in similar ways to Eq. \eqref{eq:2} and \eqref{eq:3}, respectively. Finally, the resulting features vector, say $B''(I)$, contains the background features.

\subsection{Hybrid features extraction}
\label{hybrid_features}
Hybrid features represent the mixed features that are achieved from the deep learning model pre-trained on hybrid (ImageNet+Places) datasets. The datasets consist of combined images of objects and scenes under $1,365$ categories. For the extraction of such features, we also adopt the GAP features extracted from the $p\_5$ layer (Eq. \eqref{eq:5}). Here, $H_j^i$ denotes the $i^{th}$ activation value of the corresponding $j^{th}$ feature map. $H(I)$ represents the GAP features extracted from $H$, where $H$ is the VGG-16 pre-trained model on the hybrid datasets. 
\begin{equation}
{H(I)}=\frac{1}{h*w}*\bigg\{\sum_{i=1}^{h*w}{H_{1}^i}, \sum_{i=1}^{h*w}{H_{2}^i}, ,\cdots \sum_{i=1}^{h*w}{H_{d}^i}\bigg\}.
\label{eq:5}    
\end{equation}

We again encode and normalize such a GAP features vector, which is represented by $\{h_1, h_2, \cdots, h_d\}$, extracted from Eq. \eqref{eq:5} using  Eqs. \eqref{eq:2} and \eqref{eq:3}, respectively. This produces the hybrid features, say $H''(I)$.

\subsection{Aggregation}
\label{aggregation}
These three types of deep features are aggregated to achieve our proposed features for the scene image representation. There are various simple yet efficient aggregation methods including $Min$, $Max$, $Mean$, and $Concat$. To this end, it is necessary to first find out the best and suitable aggregation method. For the selection of best aggregation method, we perform ablative analysis on various aggregation methods in Section 
\ref{aggre_methods}. It shows that the feature size becomes $512$-D for the Min, Max, and Mean methods, whereas it becomes $1,536$-D for the $Concat$ method. We choose to use the $Concat$ method on both datasets because it helps to represent three different types of information more accurately than other three methods and enable the state-of-the-art classification performance. Mathematically,
the aggregation of the three different deep features including $F''(I)$, $B''(I)$, and $H''(I)$ is shown in Eq. \eqref{eq:6}.
\begin{equation}
P(I)=\left[F''(I), B''(I), H''(I)\right],
\label{eq:6}    
\end{equation}

Alg. \ref{algo:0} lists the procedure to compute our features for the scene images representation.

\begin{algorithm}[t]
 \caption{Extraction of the proposed features}
 \begin{algorithmic}[1]
 \renewcommand{\algorithmicrequire}{\textbf{Input:}}
 \renewcommand{\algorithmicensure}{\textbf{Output:}}
 \REQUIRE $F\leftarrow$VGG-16 pre-trained on ImageNet database,\\ $B\leftarrow$VGG-16 pre-trained on Places database,\\
 $H\leftarrow$VGG-16 pre-trained on Hybrid database,\\
  $I\leftarrow$Input image for the feature extraction.
 \ENSURE  $P(I)\leftarrow$ \{\} \algorithmiccomment{Proposed features of image $I$}.\\
 \STATE Extract foreground features ($F''(I)$) of $I$ using Section \ref{foreground_features}.
 \STATE Extract background features ($B''(I)$) of $I$ using Section \ref{background_features}.
 \STATE Extract hybrid features ($H''(I)$) of $I$ using Section \ref{hybrid_features}.
 \STATE Perform aggregation to achieve $P(I)$ using Section \ref{aggregation}.
 \RETURN $P(I)$
 \end{algorithmic}
 \label{algo:0} 
 \end{algorithm}

\section{Experiment and analysis}
\label{experiment}
In this section, we discuss the experimental settings and compare our method with the previous methods, and perform ablative analysis of various parameters/components in the proposed method.

\begin{figure}[b]
\begin{center}
 \includegraphics[width=0.75\textwidth, height=50mm,keepaspectratio]{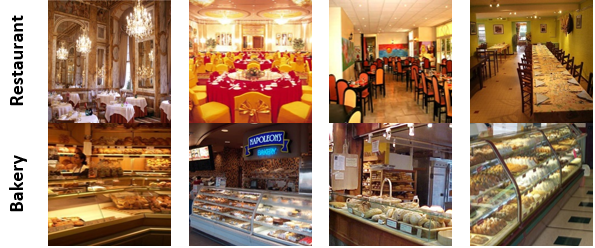}
  \caption{Example images sampled from MIT-67 \citep{quattoni_recognizing_2009}. }
  \label{fig:mit_67}
 \end{center}
  \end{figure}
  
  \begin{figure}[tb]
\begin{center}
 \includegraphics[width=0.75\textwidth, height=50mm,keepaspectratio]{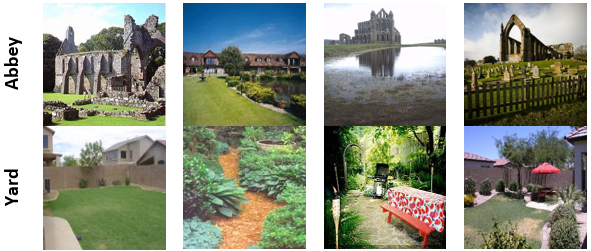}
  \caption{Example images sampled from SUN-397 \citep{xiao2010sun}. }
  \label{fig:sun_397}
 \end{center}
  \end{figure}
  
\subsection{Datasets}
For the experiments, we use two commonly used benchmark scene image datasets: MIT-67 \citep{quattoni_recognizing_2009} and SUN-397 \citep{xiao2010sun}. Both datasets contain numerous challenging images, involving within-class variations and between-class similarities which increase challenges for the classification performance. 

{\bf MIT-67} contains $15,620$ images under $67$ categories. Each category includes at least $100$ images. 
Some example images of this dataset are shown in Fig. \ref{fig:mit_67}.
For the training and testing splits, we use the same split defined by \citet{quattoni_recognizing_2009}, which has been frequently used by the existing research \citep{quattoni_recognizing_2009,
zhu_large_2010,li2010object,parizi2012reconfigurable,
juneja2013blocks,
margolin2014otc,
lin_learning_2014,
zhang2017image,
gong_multi-scale_2014,
zhou2016places,
he2016deep,
guo2016bag,
8085139,
tang_g-ms2f:_2017,
bai2019coordinate,
kim2014convolutional,
wang2019task,sitaula2019tag} 
. In particular, $80$ images per category are used as training and $20$ images per category are used as the testing in the experiments, which is a standard split defined by  \citet{quattoni_recognizing_2009}.

{\bf SUN-397} contains $108,754$ images under $397$ categories, where each category involves at least $100$ images.
Some example images of this dataset are shown in Fig. \ref{fig:sun_397}.
For training and testing, we use exactly the same splits defined by \citet{xiao2010sun}, which consists of $10$ sets of train/test splits for the experiments. Several research works 
\citep{xiao2010sun,margolin2014otc,
sanchez2013image,
gong_multi-scale_2014,
zhou2014learning,
simonyan2014very,
yang2015multi,
wu2015harvesting,
dixit2015scene,
guo2016bag,
8085139,bai2019coordinate} have been tested on this dataset using content-based features extraction methods.
To report the accuracy on this dataset, the mean accuracy of $10$ splits is used, similar to previous research. In each split defined by \citet{xiao2010sun}, $50$ images per category are used for training and $50$ images per category are used for testing. 
The total number of sampled images used in the experiments for all $10$ sets is $397,000$ ($10 \times 39,700$).

\subsection{Implementation}
To implement our method, we use Keras \citep{chollet2015keras} in Python \citep{pythoncite} language. Keras is used to implement pre-trained deep learning models \citep{simonyan2014very,gkallia2017keras_places365}) to extract foreground, background and hybrid information of the scene images. For the classification purpose, we use the $L2$-Regularized Logistic Regression ($LR$) Classifier under the LibLinear \citep{fan2008liblinear} package in Python. To tune the best cost parameters ($C$) automatically, we perform grid search in the range $\{1, 2, \cdots 50 \}$ while keeping other parameters as default. All the experiments are conducted on a laptop with a NVIDIA GeForce GTX 1050 GPU.

\subsection{Comparison with state-of-the-art methods}
We report the accuracies in two different tables: Table \ref{tab:mit67} for MIT-67 and Table \ref{tab:sun397} for SUN-397.

Table \ref{tab:mit67} reports the accuracies of the previous methods and our method on the MIT-67 dataset. To minimize the bias, we only report the accuracies of such methods that are published using such datasets.  
We also report the method types (content-based and context-based) in the tables.
For the methods belonging to content-based and context-based methods, we chose two methods (earliest one, and the latest one) for the comparison. We notice that our method produces
56.2\% higher classification accuracy than the ROI with GIST \citep{quattoni_recognizing_2009} and 1.8\% higher than the most recent method (CNN-LSTM \citep{bai2019coordinate} method under the content-based features extraction methods).
Furthermore, our method is 
29.9\% higher than the BoVW method and 5.9\% higher than the TSF \citep{sitaula2019tag} under the context-based features extraction methods. It shows that our method provides a significantly higher accuracy (\textbf{82.3}\%) on this dataset outperforming both types of methods (content and context features extraction methods). 
\begin{table}
\caption{Comparison of our proposed method using classification accuracy (\%) with the previous methods on the MIT-67 dataset. Best results are bold face.}
\vspace{0mm}
\centering
\begin{tabular}{p{8cm} p{3cm}}
\toprule
Method & Accuracy(\%)\\
\midrule
Content-based feature extraction methods\\
\midrule
ROI with GIST \citep{quattoni_recognizing_2009} &26.1 \\
MM-background \citep{zhu_large_2010} & 28.3\\
Object Bank \citep{li2010object}&37.6\\
RBoW \citep{parizi2012reconfigurable}&37.9\\
BOP \citep{juneja2013blocks}&46.1\\
OTC \citep{margolin2014otc}&47.3\\
ISPR \citep{lin_learning_2014}&50.1\\
EISR \citep{zhang2017image} &66.2\\
CNN-MOP \citep{gong_multi-scale_2014} & 68.0\\
VGG \citep{zhou2016places} & 75.3 \\
ResNet152 \citep{he2016deep} & 77.4\\
S-BoSP-fusion \citep{guo2016bag} &77.9\\
BoSP-Pre\_gp \citep{8085139}) & 78.2 \\
G-MS2F \citep{tang_g-ms2f:_2017}& 79.6\\
CNN-LSTM \citep{bai2019coordinate}& 80.5\\
\midrule
Context-based feature extraction methods\\
\midrule
BoW \citep{wang2019task} &52.5\\
CNN \citep{kim2014convolutional} & 52.0\\
s-CNN(max) \citep{wang2019task} & 54.6\\
s-CNN(avg) \citep{wang2019task} & 55.1\\
s-CNNC(max) \citep{wang2019task} & 55.9\\
TSF \citep{sitaula2019tag}& 76.5\\
\hline
\textbf{Ours} &\textbf{82.3} \\
\hline
\end{tabular}
\label{tab:mit67}
\end{table}

Table \ref{tab:sun397} presents the published accuracies of the previous methods as well as ours on the SUN-397 dataset. To date, there are not any context-based features extraction methods performed on this dataset. This may be because of the huge amount of images in this dataset that require heavy computation while performing query search on the web to achieve the context-based features. So we compare our method with some recent existing methods which belong to the content-based feature extraction methods. We observe that our method is 
28\% higher than \citet{xiao2010sun}, which is the very first method performed on this dataset, and 3.3\% higher than the recent deep learning-based method, S-BoSP-Pre\_gp \citep{8085139}. This apparently discloses the efficacy of our method which produces a noticeably higher accuracy (\textbf{66.3\%}) on this huge benchmark dataset.

\begin{table}[t]
\caption{Comparison of our proposed method using classification accuracy (\%) with the previous methods on the SUN-397 dataset.}
\vspace{0mm}
\centering
\begin{tabular}{p{7cm} p{3cm}}
\toprule
Method & Accuracy(\%)\\
\midrule
Content-based feature extraction methods\\
\midrule
Xiao et al. \citep{xiao2010sun} &38.0 \\
OTC \citep{margolin2014otc}&49.6\\
FV \citep{sanchez2013image} &47.2\\
CNN-MOP \citep{gong_multi-scale_2014} &51.9 \\
Places-CNN (\citep{zhou2014learning} &54.3 \\
Hybrid-CNN \citep{zhou2014learning} &53.8 \\
VGG-Net \citep{simonyan2014very} &51.9\\
VGG-Net-DAG \citep{yang2015multi} & 56.2\\
Metaforeground-CNN \citep{wu2015harvesting}&58.1\\
SFV-Places \citep{dixit2015scene}&61.7\\
S-BoSP-fusion \citep{guo2016bag} &62.9\\
S-BoSP-Pre\_gp \citep{8085139} & 63.7 \\
CNN-LSTM \citep{bai2019coordinate}&63.0\\
\hline
\textbf{Ours} &\textbf{66.3} \\
\hline
\end{tabular}
\label{tab:sun397}
\end{table} 

While seeing the classification accuracies of our method on both datasets (MIT-67 and SUN-397), we notice that our method produces competing and stable performance. Furthermore, regarding the feature size, our method has a lower dimensional size, which is just $1,536$-D; however, the main contender of our method (S-BoSP-Pre\_gp) on the SUN-397 dataset has $9,216$-D features size, which is over $6$ times larger than ours. Similarly, the feature size of CNN-LSTM which is the main contender on the MIT-67 dataset, still has a greater feature size ($4,096$-D) than ours. In general, the classification time will increase with a growing size of features \citep{sitaula2019unsupervised}. Therefore, our method consumes a lower classification time than such contender methods.

\subsection{Ablative study of pooling methods}
\label{pooling_methods}
\begin{figure}[t]
\begin{center}
 \includegraphics[width=\textwidth,height=70mm,keepaspectratio]{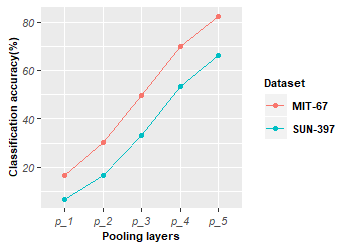}
  \caption{Comparative analysis of five pooling layers starting from $p\_1$ to $p\_5$ on MIT-67 and SUN-397 datasets. }
  \label{line_graph:pooling}
 \end{center}
  \end{figure}

The selection of appropriate pooling layer is essential in features extraction while using pre-trained deep learning models. We perform extensive experiments on both datasets using three pre-trained VGG-16 models. Specifically, we analyze five pooling layers of each of the VGG-16 models that has been trained on ImageNet, Places, and mixed data (ImageNet+Places). The five pooling layers used are $block1\_pool$, $block2\_pool$, $block3\_pool$, $block4\_pool$ and $block\_pool$, which are represented by $p\_1$, $p\_2$, $p\_3$, $p\_4$, and $p\_5$, respectively. The experimental results performed on both datasets are shown in Fig. \ref{line_graph:pooling}. In the figure, we notice that the appropriate pooling layer for the distinguishing features extraction is the $5^{th}$ pooling layer ($p\_5$). The classification accuracy using such a layer on both datasets is higher than other layers. Thus, we deduce that the $5^{th}$ pooling layer extracts the high-level information of the image that has the ability for better representation of scene images. This leads us to utilizing this layer to achieve the corresponding information including foreground, background and hybrid information of the scene images.

  \begin{figure}[tb]
\begin{center}
 \includegraphics[width=\textwidth,height=70mm,keepaspectratio]{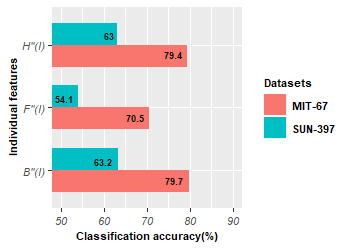}
  \caption{Comparative analysis of each individual features on the MIT-67 and SUN-397 datasets. }
  \label{bar_graph:indiv_features}
 \end{center}
  \end{figure}
  
\subsection{Ablative study of individual features}
\label{indiv_features}
We study the contribution of individual features used in our method on both datasets to see how they affect classification individually. Additionally, this study helps us to learn the highly promising features type among three different features in the scene image representation. The individual features include foreground features ($F''(I)$), background features ($B''(I)$) and hybrid features ($H''(I)$). The classification accuracies using three individual features achieved on both datasets are illustrated in Fig. \ref{bar_graph:indiv_features}. While observing the bar graph, features based on background information (background features) are better than the remaining two different types of features for the classification. Also, the superior accuracy of hybrid features than foreground features further unveil that hybrid features are also more important than foreground features on both datasets. Thus, we believe that the majority of the separability capability is attributed to the background information in most cases.

\subsection{Ablative study of aggregation methods}
\label{aggre_methods}
\begin{figure}[tb]
\begin{center}
 \includegraphics[width=\textwidth,height=70mm,keepaspectratio]{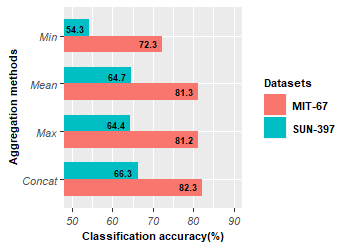}
  \caption{Comparative analysis of four aggregation methods on the MIT-67 and SUN-397 datasets. }
  \label{bar_graph:aggregation}
 \end{center}
  \end{figure}

Features aggregation is also one of the important steps in our method. For this, we perform experiments using four different aggregation methods including Minimum ($Min$), Maximum ($Max$), $Mean$ and Concatenation ($Concat$) on both datasets. The experimental result are shown in Fig. \ref{bar_graph:aggregation}. Results reveal that the $Concat$ method outperforms all other methods on both datasets. We speculate that the $Concat$ method on uniform sized features alleviates bias during features aggregation, thereby preserving multi-information uniformly for the classification purpose. As such, we adopt this method in our work.

\subsection{Ablative study of combined features}
\label{combined_features}
In this subsection, we analyze the efficacy of the combined features on both datasets using classification accuracy.
For this, we combine three types of information ($F''(I)$, $B''(I)$, and $H''(I)$) and provides four total combinations including $\left[F''(I), B''(I)\right]$, $\left[F''(I), H''(I)\right]$, $\left[B''(I), H''(I)\right]$, and $\left[F''(I), B''(I), H''(I)\right]$. For the combination of features, we use the $Concat$ aggregation method as suggested in Sec. \ref{bar_graph:aggregation} and present the results in Table \ref{tab:combined_features}. We see that the fusion of three types of features outperforms all other combinations on both datasets, in terms accuracy. We conjecture that these three types of features are complementary so that the fusion empowers a better representation of scene images. 

\begin{table}[t] 
\caption{Comparative analysis (\%) of the combined features on two datasets.}
\centering
\begin{tabular}{p{4cm} p{3cm} p{3cm}}
\toprule
Comb. layers & MIT-67 (\%) & SUN-397 (\%)\\
\midrule
$\left[F''(I),B''(I)\right]$  & 81.7&65.1 \\
$\left[F''(I),H''(I)\right]$ &80.2 &63.6\\
$\left[B''(I),H''(I)\right]$ &81.5 &65.8\\
$\left[F''(I),B''(I),H''(I)\right]$ &\bf 82.3 &\bf 66.3\\
\hline
\end{tabular}
\label{tab:combined_features}
\end{table}

\begin{table}[t] 
\caption{Computation time (seconds) consumed by three main steps such as feature extraction step, training step, and testing step for our proposed method on two datasets.}
\centering
\begin{tabular}{p{2cm} p{3cm} p{2cm} p{2cm}}
\toprule
Dataset & Feat. extraction& Training & Testing \\
& step & step & step \\
\midrule
MIT-67  & 756.7&7.8 &0.8\\
SUN-397 &4779.4 &67.8 &8.1\\
\hline
\end{tabular}
\label{tab:computation_time}
\end{table}

\subsection{Computation time}
We analyze the computation time (seconds) and list the results in Table \ref{tab:computation_time}. For SUN-397, we average the computation time of all $10$ sets. We observe that the average features extraction time per image for all the images including training and testing sets on both SUN-397 (39,7000 images) 
and  MIT-67 (6,700 images) is 0.1 seconds. Similarly, the average classification time of the testing images on the SUN-397 (19,850 images) and MIT-67 (1,340 images) are 0.0004 seconds and 0.0005 seconds, respectively.

\section{Conclusion and future works}
\label{conclusion}
In this paper, we have proposed a method that aggregates three different types of deep features for the scene image representation. Experimental results on the commonly-used benchmark scene datasets demonstrate a better classification performance of our method than the state-of-the-art methods. 
Furthermore, our method also outputs a noticeably lower size of features of the scene images. 

Our proposed method is more suitable for scene images than other types of images, because all our captured information is focused on scene images rather than images such as satellite images, biomedical images, Internet of Things (IoT) images. Those images may require other discriminating clues such as texture, global layout, temporal information, spatial information, and so on for representation. In the future, we would like to investigate other types of images for better representations.

\bibliography{mybibfile}
\end{document}